\documentclass[runningheads]{llncs}
\usepackage[T1]{fontenc}
\usepackage{graphicx}
\usepackage{booktabs}
\usepackage[misc]{ifsym}
\newcommand{\corr}{(\Letter)}
\usepackage{mwe}

\usepackage{orcidlink}
\usepackage[utf8]{inputenc} 
\usepackage{hyperref}       
\usepackage{url}            
\usepackage{amsfonts}       
\usepackage{nicefrac}       
\usepackage{microtype}      
\usepackage{xcolor}         
\usepackage[ruled,vlined,linesnumbered,commentsnumbered]{algorithm2e}

\usepackage{microtype}
\usepackage{graphicx}
\usepackage{multirow}
\usepackage{subfig}
\usepackage{booktabs} 
\usepackage{amsmath}
\usepackage{amssymb}
\usepackage{mathtools}

\usepackage{amsthm}
\usepackage{enumitem}
\usepackage[capitalize,noabbrev]{cleveref}
\usepackage{verbatim}
\usepackage{comment}
\usepackage{stfloats}
\usepackage{tikz}

\definecolor{MyDarkBlue}{rgb}{0,0.5,1}
\definecolor{MyDarkGreen}{rgb}{0.02,0.6,0.02}
\definecolor{MyDarkRed}{rgb}{0.8,0.02,0.02}
\definecolor{MyDarkOrange}{rgb}{0.40,0.2,0.02}
\definecolor{MyPurple}{RGB}{111,0,255}
\definecolor{MyRed}{rgb}{1.0,0.0,0.0}
\definecolor{MyGold}{rgb}{0.75,0.6,0.12}
\definecolor{MyDarkgray}{rgb}{0.66, 0.66, 0.66}

\newcommand{\model}[1]{{life-long world model}}
\newcommand{\replay}[1]{{experience replay mechanism}}
\newcommand{\newcontrib}[1]{{exploratory-conservative behavior learning}}

\usepackage{pifont}

\newcommand{\yes}{\scriptsize\ding{51}} 
\newcommand{\no}{\scriptsize\ding{55}}  
\newcommand{\x}{o}
\newcommand{\z}{z}
\newcommand{\M}{{\mathcal{M}}}
\newcommand{\G}{{\mathcal{G}}}

\makeatother

\newcommand{\eqn}[1]{Eq.~\eqref{#1}}
\newcommand{\sect}[1]{Section~\ref{#1}}

\newcommand{\fig}[1]{Fig.~\ref{#1}}

\begin{document}

\title{Continual Visual Reinforcement Learning with\\A Life-Long World Model}



\author{Minting Pan 
\and Wendong Zhang  
\and Geng Chen
\and Xiangming Zhu
\and Siyu Gao \and \\ 
Yunbo Wang \corr \and
Xiaokang Yang}


\authorrunning{M. Pan et al.}

\institute{MoE Key Lab of Artificial Intelligence, AI Institute, Shanghai Jiao Tong University \email{\{panmt53, diergent, yunbow, xkyang\}@sjtu.edu.cn}}

\tocauthor{Minting Pan, Wendong Zhang, Geng Chen, Xiangming Zhu, Siyu Gao, Yunbo Wang, Xiaokang Yang}
\toctitle{Continual Visual Reinforcement Learning with A Life-Long World Model}

\maketitle              

\begin{abstract}

Learning physical dynamics in a series of non-stationary environments is a challenging but essential task for model-based reinforcement learning (MBRL) with visual inputs. It requires the agent to consistently adapt to novel tasks without forgetting previous knowledge. In this paper, we present a new continual learning approach for visual dynamics modeling and explore its efficacy in visual control. The key assumption is that an ideal world model can provide a non-forgetting environment simulator, which enables the agent to optimize the policy in a multi-task learning manner based on the imagined trajectories from the world model. To this end, we first introduce the \textit{life-long world model}, which learns task-specific latent dynamics using a mixture of Gaussians and incorporates generative experience replay to mitigate catastrophic forgetting. Then, we further address the value estimation challenge for previous tasks with the \textit{exploratory-conservative behavior learning} approach. Our model remarkably outperforms the straightforward combinations of existing continual learning and visual RL algorithms on DeepMind Control Suite and Meta-World benchmarks with continual visual control tasks. Code available at \url{https://github.com/WendongZh/continual_visual_control}.

\end{abstract}

\section{Introduction}
\label{sect:triplet_shift}

\begin{figure}[t]
    \includegraphics[width=\textwidth]{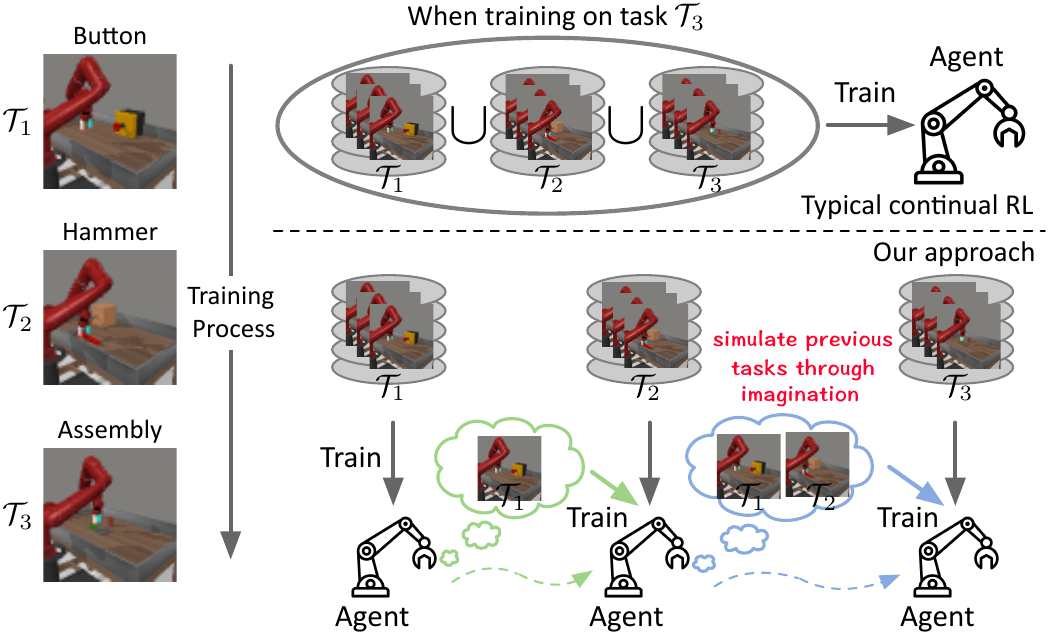}
    \caption{\textbf{Left:} The problem setting of continual visual control.
    \textbf{Top Right:} Existing continual RL solutions typically retain large amounts of previous data in a replay buffer to train model-free RL models. 
    \textbf{Bottom Right:} We employ model-based RL to alleviate catastrophic forgetting in continual visual control. The key insight is to simulate past tasks using a lifelong world model (LLWM). 
    }
    \label{fig:intro}
\end{figure}

Recent advances in reinforcement learning (RL) have demonstrated remarkable success in mastering complex visual environments through world models~\cite{ha2018recurrent,hafner2020dream,paniso}. 
However, these approaches typically assume a static task distribution and fixed environmental dynamics, which fundamentally break down in real-world scenarios where agents must adapt to continuously evolving conditions. 
In domains such as kitchen robotics, autonomous systems, or dynamic clinical environments, as shown in \fig{fig:intro}, agents are required to learn incrementally from non-stationary data streams while retaining and building upon prior knowledge—a capability central to continual learning.

A critical challenge in this setting is catastrophic forgetting, where neural networks overwrite previously acquired skills when trained on new tasks. 
This issue is exacerbated in visual RL, where agents must process pixel-based observations that encode both task-specific and task-agnostic features. 
Unlike the previous efforts to integrate standard solutions for catastrophic forgetting into model-free RL methods~\cite{DBLP:conf/nips/WolczykZPKM21,xie2022lifelong,powers2022cora,zhou2022forgetting}, it is impractical to retain a large data buffer for each previous task.
In this paper, we suggest that the key challenge is rooted in combating the distribution shift in dynamics modeling. 
The basic assumption of our approach is that \emph{an ideal world model can be viewed as a natural remedy against catastrophic forgetting in continual RL scenarios, because it can replay the previous environments through ``imagination'', allowing multi-task policy optimization.}

To cope with the dynamics shift, we introduce a novel life-long world model (LLWM) architecture, which lies a dynamically expandable state representation space governed by task index in the lifelong learning.
Specially, it learns a mixture of Gaussian priors to capture task-specific latent dynamics based on a set of categorical task variables. 
Building upon LLWM, we formulate an experience replay mechanism to synthesize fictitious trajectories of previous tasks, avoiding huge raw pixel observation storage.
To achieve that, we need to train an additional generative model to reproduce the initial video frames in previous tasks and then feed them into the learned LLWM to generate subsequent image sequences for data rehearsal.
The training process alternates between (i) \textit{generating rehearsal data with the frozen world model learned on previous tasks} and (ii) \textit{training the entire model with both rehearsal data and current real data}.

However, a straightforward use of the experience replay mechanism is less effective.
In world model learning, the mechanism's replay of previous task trajectories suffers from excessive homogeneity, inducing an overfitting issue in the reward prediction that may overestimate or underestimate the true reward. 
Moreover, as shown in \fig{fig:method_exp1}, the value network increasingly overestimates the true value functions of previous tasks during behavior learning as the number of sequential tasks grows. This issue is fundamentally rooted in the constrained diversity of replayed data.
To solve these issues, we propose the \textit{\newcontrib} approach, which improves the aforementioned solution in two aspects. 
First, in representation learning, we use the $\epsilon$-greedy exploration and introduce an action-shuffling technique to augment the replayed observation-action-reward trajectories, thereby preventing the reward predictor of the world model from overfitting the narrow distribution of replayed data.
Second, to alleviate value overestimation in behavior learning, we use the value network learned in the previous task to constrain the target of value estimation on replayed data.
This learning strategy balances the exploration and constraints of value functions in historical tasks, \textit{i.e.}, it encourages the agent to learn a conservative policy over a broader distribution of observation-action-reward pairs.

We evaluate our approach on DeepMind Control~\cite{tassa2018deepmind} and Meta-World~\cite{DBLP:conf/corl/YuQHJHFL19} environments.
For the DeepMind Control, we construct sequential tasks with different robotic physical or environmental properties. 
For the Meta-World, we collect various tasks with distinct control policies and spatiotemporal data patterns.
Experiments on these two benchmarks show that our approach remarkably outperforms the na\"ive combinations of existing continual learning algorithms and visual MBRL models. 

In summary, the main technical contributions of this paper are as follows:
\begin{itemize}[leftmargin=*]
\vspace{-5pt}
    \item 
    We present a novel architecture of \textit{life-long world model} within the MBRL framework, which learns task-specific Gaussian priors and performs generative replay to mitigate catastrophic forgetting.
    \vspace{2pt}\item 
    We propose the \textit{\newcontrib{} method} to mitigate reward overfitting and value overestimation issues in continual RL. 
\end{itemize}

\section{Related Work}
 
\noindent \textbf{Continual RL.} 
Continual RL has become a hot topic in recent years~\cite{khetarpal2022towards,hadsell2020embracing,DBLP:conf/nips/WolczykZPKM21,powers2022cora}. 
To overcome catastrophic forgetting, a straightforward way is to explicitly retain pre-learned knowledge, which includes storage-based approaches that directly save task-dependent parameters~\cite{ammar2014online} or priors~\cite{shi2021meta,yu2020gradient,liu2021conflict}, distillation based approaches~\cite{zhang2022catastrophic,lan2022memory,igl2021transient} that leverages knowledge distillation~\cite{hinton2015distilling} to recall previous knowledge, and the rehearsal based approaches that simply save trajectories~\cite{daniels2022model,liotet2022lifelong} or compact representations~\cite{riemer2019scalable} for re-training.
Besides, previous approaches~\cite{zhou2022forgetting,xie2022lifelong} mainly consider Markov decision processes (MDP) with compact state space, which leaves the control tasks with high-dimensional visual inputs unexplored. 
Inspired by the success of continual visual forecasting~\cite{chen2022continual}, our work proposes a new generative framework to tackle the forgetting issues in continual visual control tasks.
Instead of storing extensive trajectories, our method preserves only a single set of model parameters and leverages the world model with an experience replay mechanism to retain past knowledge, ensuring a more efficient and effective approach.
Besides knowledge retention, there are also approaches using shared structures to learn non-forgetting RL agents~\cite{mendez2022modular,mendez2022reuse}.

\vspace{5pt}
\noindent \textbf{RL for visual control.}
In visual control tasks, agents can only access high-dimensional observations.
Previous approaches can be roughly summarized into two categories, that is, model-free approaches~\cite{yarats2021improving,DBLP:conf/icml/LaskinSA20,kostrikov2020image} and model-based approaches~\cite{hafner2019learning,hansen2022modem,sekar2020planning}.
Compared with MFRL approaches, the MBRL approaches explicitly learn the dynamic transitions and generally obtain higher sample efficiency.
Ha and Schmidhuber proposed the World Model~\cite{ha2018recurrent} with a two-stage training procedure that first learns latent transitions of the environment with self-supervision and then performs behavior learning on the states generated by the world model.
PlaNet~\cite{hafner2019learning} introduces the\textit{ recurrent state-space model} (RSSM) to optimize the policy over its recurrent states.
The same architecture is also adopted by DreamerV1-V3~\cite{hafner2020dream,hafnermastering,hafner2023mastering}, where behavior learning is conducted on the latent imagination of RSSM.
Recently, several studies~\cite{DBLP:conf/iclr/BharadhwajBEL22,paniso,sun2024learning,burchi2025learning,li2025open}, such as Iso-Dream~\cite{paniso} and TWISTER~\cite{burchi2025learning}, further have explored different ways to produce more robust dynamics representations for MBRL.
In this paper, we make three improvements to adapt MBRL to continual visual control, \textit{i.e.}, the world model architecture, the world model learning scheme, and behavior learning.

\section{Approach}

Given a stream of tasks $\{\mathcal{T}_1, \ldots, \mathcal{T}_K\}$, we formulate continual visual control as a partially observable Markov decision process (POMDP) in sequential domains. 
At a particular task $\mathcal{T}_{k}$ where $k$ represents the task index, the POMDP contains high-dimensional visual observations and scalar rewards $({o}_{t}^{k}, r_{t}^{k}) \sim p({o}_{t}^{k}, r_{t}^{k} | {o}_{\textless t}^{k}, a_{\textless t}^{k})$ that are provided by the environment, and continuous actions $a_{t}^{k} \sim p(a_{t}^{k} | o_{\leq t}^{k}, a_{\textless t}^{k})$ that are drawn from the action model. 
The goal is to maximize the expected total payoff on all tasks $\mathbb E_\pi(\sum_k \sum_t r_{t}^{k})$.

Compared to previous RL approaches, we aim to build an agent that can learn the state transitions in different tasks, such that
\begin{equation}
\begin{split}
\label{eq:setup_control}
    \hat{o}_{t}, \hat{z}_{t}, \hat{r}_{t} \sim \mathcal{M}({o}_{t-1}, a_{t-1}, {k}), \quad
    {a}_{t} \sim \mathcal{\pi}(\hat{z}_{t}, {k}),
\end{split}
\end{equation}
where $\mathcal{M}$ and $\mathcal{\pi}$ are respectively the world model and the action model, and $\hat{o}_{t}$, $\hat{z}_{t}$ and $\hat{r}_{t}$ are respectively the generated next observation, the extracted latent state, and the predicted reward.

In this section, we present the details of our approach, which mainly consist of two parts: \textit{life-long world model} and \textit{exploratory-conservative behavior learning}.
In~\sect{sect:pipeline}, we first discuss the overall pipeline of our method and its basic motivations.
In \sect{sect:mixturewm} to \sect{sect:dul_reg}, we respectively introduce the technical details of the two components.

\subsection{Overall Pipeline}
\label{sect:pipeline}

Our overall pipeline is built upon DreamerV2 method~\cite{hafnermastering}, which involves two iterative stages: (i) world model learning of latent state transitions and reward predictions and (ii) behavior learning via latent imagination.
To adapt DreamerV2 to the continual RL setup, we make improvements in three aspects: the world model architecture, the world model learning scheme, and behavior learning.
As shown in Fig.~\ref{fig:intro}, we first initialize the agent, including the life-long world model $\mathcal{M}$, the actor $\pi$, the critic $v$, and an image generator $\mathcal{G}$. 
Since the agent is not allowed to access real data of previous tasks, we will retain a copy of the current models after finishing the training process of a particular task, termed as $\mathcal{M}^{\prime}$, ${\pi}^{\prime}$, ${v}^{\prime}$, and $\mathcal{G^{\prime}}$. 
These frozen models generate replay trajectories for previous tasks through imagined rollouts using images synthesized by $\mathcal{G^{\prime}}$.
In other words, we train the models on sequential tasks but continually update one copy of them to involve increasingly more previous knowledge for rehearsal.
This pipeline is at the heart of our approach. For the control tasks, ideally reproducing prior knowledge as latent imagination is a cornerstone for alleviating the forgetting issues in behavior learning.
Besides, the requirement of ideally reproducing prior knowledge also guides the detailed world model design.

\subsection{Life-long World Model}
\label{sect:mixturewm}
 
The proposed \textit{\model{}} (LLWM) incorporates Gaussian mixture representations. 
We improve the world model $\M$ in DreamerV2~\cite{hafnermastering} from the perspective of multimodal spatiotemporal modeling, which serves as the cornerstone to overcome catastrophic forgetting in continual visual control.

As discussed in \sect{sect:triplet_shift}, the key challenge for the world model lies in how to be aware of the distribution shift.
To address this challenge, we exploit mixture-of-Gaussian variables to capture the multimodal distribution of both visual dynamics in the latent space and spatial appearance in the input/output observation space.
Accordingly, the world model can be written as:
\begin{equation}
\begin{split}
    \text{Representation module:} & \quad \z_{t} \sim q_{\phi}({\x}_{t}^k, \z_{t-1}, a_{t-1}^{k},  k) \\
    \text{Transition module:} & \quad \hat{\z}_{t} \sim p_{\psi}(\z_{t-1}, a_{t-1}^{k}, k) \\
    \text{Observation module:} & \quad \hat{\x}_{t} =  p_{{\theta}_{1}}(\z_{t}, k) \\
    \text{Reward module:} & \quad \hat{r}_{t} =  p_{{\theta}_{2}}(\z_{t}, k). \\
\end{split}
\label{eq:Cdreamer}
\end{equation}
The representation module aims to encode observations and actions to infer the latent state $\z_{t}$. It also takes the categorical task variable $k \in \{1,\ldots,K\}$ as an extra input to handle the covariate shift in input space.
The transition module is also aware of the categorical task variable and tries to predict latent state $\hat{\z}_{t}$ to approximate the posterior state $\z_{t}$. These modules are jointly optimized with the Kullback-Leibler divergence to learn the posterior and prior distributions of $\z_{t}$.
The observation and reward modules take as inputs $\z_{t}$ along with the task variable $k$ to reconstruct images and predict corresponding rewards.

In this model, the task-specific latent variables $\z_{t}$ and $\hat{\z}_{t}$ have the form of Gaussian mixture distributions conditioned on $k$, which shares a similar motivation with the Gaussian mixture priors proposed in existing unsupervised learning methods~\cite{dilokthanakul2016deep,jiang2017variational,rao2019continual}.
Compared with these methods, LLWM leverages multimodal priors to model spatiotemporal dynamics more effectively.
For task $\mathcal T_k$, the objective function of the world model combines the reconstruction loss of the input frames, the prediction loss of the rewards, and the KL divergence:
\begin{equation}
\begin{split}
    {\mathcal{L}}_{\M} &= \mathbb{E} \big[ \sum\limits_{t=1}^{H} - \log p(\hat{\x}_{t} \, | \, \z_{t}, k) - \log p(\hat{r}_{t} \, | \, \z_{t}, k) \\ &-\alpha D_{KL}(q(\z_{t} \, | \, {\x}_{t}^k, \z_{t-1}, a_{t-1}^{k},  k) \, || \, p(\hat{\z}_{t} \, | \,  \z_{t-1}, a_{t-1}^{k}, k) \big], 
\end{split}
\label{eq:dynamic}
\end{equation}
where $\alpha$ equals to $1.0$, and $({\x}_{t}^k, a_{t}^{k}, r_{t}^{k})_{t=1}^H$ represents a training trajectory sampled from the data buffer for the current task or produced by the following \replay{} for the historical tasks.

\vspace{5pt}
\noindent \textbf{Generative experience replay.}
We train LLWM under a generative experience replay scheme, which additionally employs an image generator to produce the initial frames in the replayed trajectories of previous tasks.
To overcome the covariant shift of the image appearance in time-varying environments, the image generator also uses Gaussian mixture distributions to form the latent priors, denoted by $\boldsymbol{e}$.
Specifically, after training on the previous task, we first retain a copy of the image generator $\G$, the world model $\M$, and the action model $\pi$. These copies are respectively denoted as $\mathcal{G^{\prime}}$, $\mathcal{M^{\prime}}$, and ${\pi}^{\prime}$.
Then, for each previous task $\mathcal T_{\tilde{k}<k}$, we use $\mathcal{G^{\prime}}$ to generate the initial frames of the rehearsal trajectories. Along with a zero-initialized action $a_0$, we iteratively perform $\mathcal{M^{\prime}}$ and $\mathcal{\pi}^{\prime}$ to  generate future sequences:
\begin{equation}
    \begin{split}
        \hat{\x}^{\tilde{k}}_{0} \leftarrow \mathcal{G^{\prime}}({\tilde{k}}), \quad
        \hat{\x}_{t}^{\tilde{k}}, \hat{z}_{t}, \hat{r}_{t}^{\tilde{k}} \leftarrow \mathcal{M^{\prime}}(\hat{\x}^{\tilde{k}}_{t-1},\hat{a}_{t-1}^{\tilde{k}},{\tilde{k}}),  \quad
        \hat{a}_{t}^{\tilde{k}} \leftarrow \mathcal{\pi}^{\prime}(\hat{z}_{t}, {\tilde{k}}).
    \end{split}
\end{equation}
Finally, the replay trajectories $(\hat{\x}_{t}^{\tilde{k}}, \hat{a}_{t}^{\tilde{k}}, \hat{r}_{t}^{\tilde{k}})_{t=1}^{H}$ and the real trajectories at the current task $\mathcal T_k$ are mixed to train the generator and the agent in turn, and $\mathcal{G^{\prime}}$, $\mathcal{M^{\prime}}$, and $\mathcal{\pi}^{\prime}$ will be frozen and continually updated until the task stream ends.
%
Since the world model is also involved in the replay process and plays a key role, the proposed experience replay mechanism is significantly different from all previous generative replay methods.
During task $\mathcal{T}_k$, the world model $\M$ is trained by minimizing
\begin{equation}
\begin{split}
    {\mathcal{L}}_{\M} = \sum\limits_{\tilde{k}=1}^{k-1} {\mathcal{L}}_{\M}^{\tilde{k}}(\hat{\x}_{1:H}^{\tilde{k}},\hat{a}_{1:H}^{\tilde{k}}, \hat{r}_{1:H}^{\tilde{k}}) + {\mathcal{L}}_{\M}^k(\x_{1:H}^k,a_{1:H}^k, r_{1:H}). \\
\end{split}
\label{eq:M_loss}
\end{equation}
The objective function of the image generator $\G$ can be written as
\begin{equation}
\begin{split}
    {\mathcal L}_{\G} & = \ \mathbb{E}_{q(\boldsymbol{e} \, | \, {{\x}_{1}^k,k})} \log p({\x}_{1}^k \, | \, \boldsymbol{e}, k) - \beta D_{KL}(q(\boldsymbol{e} \, | \, {\x}_{1}^k, k) \, || \, p(\hat{\boldsymbol{e}} \, | \, k)) \\
    & + \sum\limits_{\tilde{k}=1}^{k-1} \big[ \mathbb{E}_{q(\boldsymbol{e} \, | \, {\hat{{\x}}_{1}^{\tilde{k}},{\tilde{k}}})} \log p(\hat{{\x}}_{1}^{\tilde{k}} \, | \, \boldsymbol{e},{\tilde{k}}) - \beta D_{KL}(q(\boldsymbol{e} \, | \, \hat{{\x}}_{1}^{\tilde{k}}, {\tilde{k}}) \, || \, p(\hat{\boldsymbol{e}} \, | \, {\tilde{k}})) \big],\\
\end{split}
\label{eq:G_loss}
\end{equation}
where we use the $\ell_2$ loss for reconstruction and set $\beta$ equals to $10^{-4}$ through an empirical grid search.

\subsection{Exploratory-Conservative Behavior Learning}
\label{sect:dul_reg}

\begin{figure}[t]
\centering
\includegraphics[width=\linewidth]{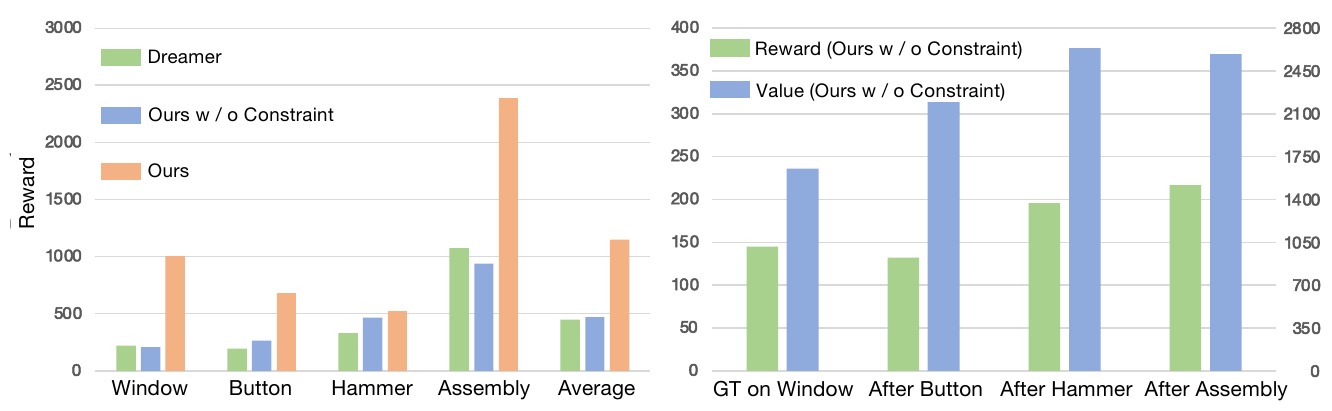}
\caption{Motivating examples. \textbf{Left:} Effect of the proposed behavior learning method. We evaluate the model obtained after the last training task (\textit{i.e.}, \textit{Assembly} on Meta-World) on each of the tasks represented on the X-axis. The task order is \textit{Window-open} $\rightarrow$ \textit{Button-press} $\rightarrow$ \textit{Hammer} $\rightarrow$ \textit{Assembly}. ``Constraint'' represents the exploratory-conservative behavior learning. 
\textbf{Right:} Reward predictions and value estimations in the na\"ive solution. We use $100$ batches of trajectories of \textit{Window-open} to evaluate models trained after each subsequent task. We use the model trained on \textit{Window-open} to calculate the ``ground truth'' of the state values. The X-axis represents different models that just finished training on corresponding tasks. The left and right Y-axes represent reward and value, respectively.
}
\label{fig:method_exp1}
\vspace{-5pt}
\end{figure}

\noindent \textbf{Preliminary findings and motivations.}
The life-long world model serves as a na\"ive solution to handle continual visual control tasks.
However, as shown in Fig~\ref{fig:method_exp1}(left), on the Meta-World benchmark, we can observe that this na\"ive solution (indicated by the blue bars) can not mitigate the forgetting issue and may even degenerate the control performance on some tasks compared with Dreamer, such as the task \textit{Window-open}.
We find that the performance degeneration is rooted in two closely related overfitting problems that happen respectively in world model learning and behavior learning. Both of them lead to the erroneous estimation of the value functions for previous tasks.
As shown in Fig.~\ref{fig:method_exp1}(right), we use the same trajectory batches from the very first task (\textit{i.e.}, \textit{Window-open}) to evaluate the outputs of the reward predictor and those of the value network after training these models on each subsequent task. 
These issues form a vicious cycle: biased reward predictions amplify value estimation errors, while overestimated values further distort policy optimization. It motivates our dual focus on enhancing the robustness of the reward model in the world model and constraining value extrapolation for stable continual learning.

\vspace{5pt}
\noindent \textbf{Solutions.}
To tackle these problems, we propose the exploratory-conservative behavior learning method (see Alg.~\ref{alg:cdreamer}), which simultaneously enhances the world model's reward prediction capability and stabilizes value estimation in actor-critic optimization. 
First, we introduce specific data augmentation for the reward predictor during experience replay.
The key insight lies in that we want the world model to experience more diverse observation-action pairs to keep it from overfitting the limited rehearsal data and thus underperforming for out-of-distribution samples.
On one hand, we exploit the $\epsilon$-greedy strategy to improve action exploration to generate trajectories of previous tasks $(\hat{\x}_{t}^k, \hat{a}_{t}^{k}, \hat{r}_{t}^{k})_{t=1}^{H}$.
On the other hand, we randomly shuffle the generated actions and reuse the learned world model at the previous task $\M^{\prime}$ to predict new rewards, which results in a new trajectory $(\hat{\x}_{t}^k, \tilde{a}_{t}^{k}, \tilde{r}_{t}^{k})_{t=1}^{H}$. 
Notably, these augmented trajectories are only used to train the reward predictor. They break the temporal biases in the distribution of rehearsal data and prevent the reward prediction from overfitting issues. 
Assuming $\tilde{\z}_{t}$ represents the latent state extracted from the augmented data, the objective function of the world model in Eq.~\eqref{eq:dynamic} can be rewritten as:
\begin{equation}
\begin{split}
    {\mathcal{L}}_{\M} &= \mathbb{E} \big[ \sum\limits_{t=1}^{H} - \log p(\hat{\x}_{t} \, | \, \z_{t}, k) - \log p(\hat{r}_{t} \, | \, \z_{t}, k) \\ &- \delta \log p(\tilde{r}_{t} \, | \, \texttt{sg}(\tilde{\z}_{t}), k) -\alpha D_{KL}(q(\z_{t} \, | \, {\x}_{t}^k, \cdot) \, || \, p(\hat{\z}_{t} \, | \,  \cdot) \big], 
\end{split}
\label{eq:dynamic_replay}
\end{equation}
where $\texttt{sg}$ is the stop-gradient operation and $\delta$ equals to $0.5$ in all experiments.

\begin{algorithm}[t] 
  \caption{Exploratory-conservative behavior learning}  
  \label{alg:cdreamer}  
  \SetAlgoLined  
  \textbf{Input: }{Task stream $\mathcal{T}_1,  \ldots, \mathcal{T}_K$}  \\
    \textbf{Output:} World model $\M$, generator $\G$, and actor $\mathcal{\pi}$  \\
    Initial buffer $\mathcal{B}_{1}$ at $\mathcal{T}_1$ with random episodes  \\
    \While{not converged}{
        Train $\G$ at $\mathcal{T}_1$ according to \eqn{eq:G_loss} with $k=1$   \\
        Perform dynamics learning at $\mathcal{T}_1$ according to \eqn{eq:dynamic}  \\
        Perform behavior learning at $\mathcal{T}_1$ according to \eqn{eq:value}  \\
        Add interaction experience to buffer $\mathcal{B}_{1}$ 
    }  
    
    \For{$k=2, \ldots, K$}{
        Initial buffer $\mathcal{B}_{k}$ at $\mathcal{T}_k$ with random episodes \\
        Retain a copy of previous model as $\mathcal{M^{\prime}}$, $\mathcal{G^{\prime}}$, $\mathcal{\pi^{\prime}}$ , and $v^{\prime}$  \\
        Generate replayed trajectory $\tau_{1}$ and augmented data $\tau_{2}$  \\
        \tcp{Mix replayed data at $\mathcal{T}_{1:k-1}$ and real data at $\mathcal{T}_k$} 
        $ \mathcal{B}_k^{\prime}  \leftarrow \tau_{1}  \cup \mathcal{B}_k$  \\
        \While{not converged}{
            Train $\G$ on $ \mathcal{B}_k^{\prime}$ according to \eqn{eq:G_loss}  \\
            Perform dynamics learning on $ \mathcal{B}_k^{\prime}$ with \eqn{eq:dynamic}  \\
            Perform behavior learning on $ \mathcal{B}_k$ with \eqn{eq:value}  \\
            \tcp{Use reward augmentation}
            Train reward module on $\tau_{2}$ with \eqn{eq:dynamic_replay}  \\
            \tcp{Use conservative value target} 
            Perform behavior learning on $\tau_{1}$ with \eqn{eq:new_value}  \\
            Append interaction experience to $\mathcal{B}_{k}$ }
    }
\end{algorithm}

Second, we constrain the target of the value model when performing behavior learning on the replayed data, which is inspired by \textit{conservative Q-learning} (CQL)~\cite{DBLP:conf/nips/KumarZTL20}.
Given the world model that captures the latent dynamics, the behavior learning stage is performed on the task-specific latent imagination.
We use $t^\prime$ to denote the time index of the imagined states.
Starting at the posterior latent state $z_{t}$ inferred from the visual observation, we exploit the transition module, the reward predictor, and the action model to predict the following states and corresponding rewards in imagination, which are all guided by the explicit task label $k$.
Then, the action model and the value model will be optimized on the imagined trajectories:
\begin{equation}
\label{eq:ac-model}
\begin{split}
    \text{Action model:}& \quad {a}_{t}^{k} \sim \pi(z_{t} , k ) \\
    \text{Value model:}& \quad v(z_{t^\prime}, k) \approx \mathbb{E}_{\pi\left(\cdot \mid z_{t^\prime}, k\right)} \sum_{t^\prime=t}^{t+L} \gamma^{t^\prime-t} r_{t^\prime},
\end{split}
\end{equation}
where $L$ is the imagination time horizon and $\gamma$ is the reward discount.
The action model is optimized to maximize the value estimation, while the value model is optimized to approximate the expected imagined rewards. 
The training target for the value model on real data is:
\begin{eqnarray}
    V_{t} = r_{t} + {\lambda}_{t}
    \begin{cases}
        (1 - \lambda) v(z_{t+1}) + \lambda V_{t+1} & if \quad t < L, \\
         v(z_{L}) & if \quad t = L,
    \end{cases}
\label{eq:value}
\end{eqnarray}
where $\lambda$ equals to $0.95$. We adopt the objective functions from DreamerV2~\cite{hafnermastering} to train these models.
However, when learning on the replayed data, this objective may result in the overestimation problem as shown in Fig.~\ref{fig:method_exp1}.
Therefore, we reuse the learned value model to update the target when training on replayed data.
Given a replayed trajectory $(\hat{\x}_{t}^k, \hat{a}_{t}^{k}, \hat{r}_{t}^{k})_{t=1}^{H}$, we first calculate the training target $V_{t}$ according to Eq.~\eqref{eq:value}.
We then retain a copy of the value model $v^{\prime}$ that was learned on the previous task and use the same trajectory as above to produce another conservative target $\widetilde{V}_{t}$ with the same function.
We use $\widetilde{V}_{t}$ to constrain the current value model to tackle the value overestimation problem for previous tasks.
The final target for training the value model on the rehearsal data is
\begin{eqnarray}
    V_{t}^{\prime} = 
    \begin{cases}
        V_{t} & if \quad V_{t} < \widetilde{V}_{t}, \\
        \widetilde{V}_{t} & if \quad V_{t} \ge \widetilde{V}_{t}.
    \end{cases}
\label{eq:new_value}
\end{eqnarray}
Alg.~\ref{alg:cdreamer} gives the full training procedure of the \newcontrib{}. With the above two improvements, it successfully extends MBRL algorithms to sequential visual control tasks.

\section{Experiments}
\label{sec:expri}

\begin{figure}[t]
\centering
\includegraphics[width=\linewidth]{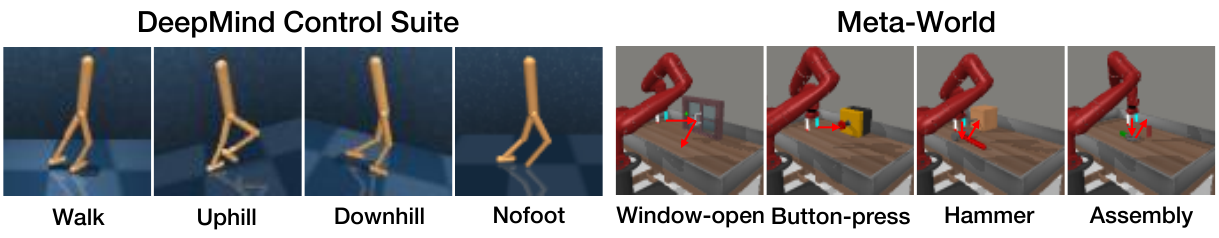}
\caption{Showcases of the sequential visual control tasks.
}
\label{fig:RL_bench}
\end{figure}

\subsection{Implementations}
\noindent \textbf{Benchmarks.}
We exploit two RL platforms with rich visual observations to perform the quantitative and qualitative evaluation for continual control tasks:
\vspace{-5pt}
\begin{itemize}[leftmargin=*]
    \vspace{5pt}\item \textbf{Continual DeepMind Control Suite}~\cite{tassa2018deepmind}. The DeepMind Control Suite (DMC) is a popular RL environment that contains various manually designed control tasks with an articulated-body simulation. We use the \textit{Walker} as the base agent and construct a task stream including four different tasks, \textit{i.e.}, \textit{Walk} $\rightarrow$ \textit{Uphill} $\rightarrow$ \textit{Downhill} $\rightarrow$ \textit{Nofoot}. \textit{Nofoot} represents the task in which we cannot control the right foot of the robot.
    \vspace{5pt}\item \textbf{Continual Meta-World}~\cite{DBLP:conf/corl/YuQHJHFL19}. The Meta-World environment contains $50$ distinct robotic manipulation tasks with the same embodied agent. We select four tasks to form a task stream for evaluation: \textit{Window-open} $\rightarrow$ \textit{Button-press} $\rightarrow$ \textit{Hammer} $\rightarrow$ \textit{Assembly}.  
\end{itemize}
We provide some showcases in Fig.~\ref{fig:RL_bench}. We assume that the task label in the test phase is provided for all experiments. We use the same task order for comparisons between different methods, and without loss of generality, our approach is practical to different task orders (see \sect{sec:ablation}).

\vspace{5pt}
\noindent \textbf{Compared methods.}
We compare our method with the following baselines:
\vspace{-2pt}
\begin{itemize}[leftmargin=*]
    \vspace{5pt}\item \textbf{DreamerV2 (DV2)}~\cite{hafnermastering}: It is an MBRL method that learns the policy directly from latent states in the world model. 
    \vspace{5pt}\item \textbf{CURL}~\cite{DBLP:conf/icml/LaskinSA20}: It is a model-free RL framework that exploits unsupervised representation learning to help high-level feature extraction from visual inputs. We use the version that uses SAC~\cite{DBLP:conf/icml/HaarnojaZAL18} for policy optimization.
    \vspace{5pt}\item \textbf{RL with EWC} \cite{kirkpatrick2017overcoming}: It is a mainstream parameter-constrained continual learning method. We apply it to both DreamerV2 and CURL. 
\end{itemize}

\noindent \textbf{Training details.} In all environments, the input image size is set to $64\times64$, the batch size is $50$, and the imagination horizon is $50$. For each task in Meta-World, we train our method for $200$K iterations with action repeat equal to $1$, which results in $200$K environment steps. For each task in DMC, we train our method for $500$K iterations with action repeat equals to $2$, which results in $1$M environment steps. The probability in the $\epsilon$-greedy exploration is set to $0.2$ for all experiments. 

\begin{table*}[t]
\centering
    \caption{Results on the DeepMind Control Suite: The task order is \textit{Walk} $\rightarrow$ \textit{Uphill} $\rightarrow$ \textit{Downhill} $\rightarrow$ \textit{Nofoot}. We evaluate the models on all tasks after completing training on the last task (\textit{Nofoot}). 
    The last two rows represent baseline models: one is a model jointly trained on all tasks, and the other is a model trained independently on a single task.
    }
    \vspace{4pt}
\small
\setlength\tabcolsep{4pt}
    \begin{tabular}{lccccc}
    \toprule
    Method & Walk & Uphill  & Downhill & Nofoot & Average  \\
    \midrule
    DV2 &239 $\pm$ 156  & 181 $\pm$ 19 & 444 $\pm $178   &936 $\pm$ 32   & 450   \\
    CURL  & 539 $\pm$ 170 & 132 $\pm$ 64  & 133 $\pm$ 63    & 919 $\pm$ 46   &  431  \\
    DV2+EWC (hyper-1) &\textbf{744} $\pm$ 36 &27 $\pm$ 12 & 70 $\pm$ 29 &49 $\pm$ 40  &223 \\
    DV2+EWC (hyper-2) &705 $\pm$ 39 &351 $\pm$ 133 & 292 $\pm$ 139 &423 $\pm$ 182 & 443 \\
    CURL+EWC (hyper-3)  & 206 $\pm$ 125 & 119 $\pm$ 90 & 186 $\pm$ 113     & 433 $\pm$ 322  & 236  \\
    CURL+EWC (hyper-4) & 222 $\pm$ 143 & 101 $\pm$ 59 & 409 $\pm$ 88     & 887 $\pm$ 75  & 405  \\
    Ours & 606 $\pm$ 365  & \textbf{734} $\pm$ 346 & \textbf{954} $\pm$ 30     & \textbf{951} $\pm$ 19 &\textbf{812} \\
    \midrule
    Multi-task  & 951 $\pm$ 16  & 581 $\pm$ 120 &  903 $\pm$ 34    & 950 $\pm$ 16 & 846 \\
    Single-task  & 759 $\pm$ 24  & 343 $\pm$ 97  & 934$\pm$ 24    & 929 $\pm$ 35 & 741 \\
    \bottomrule
    \end{tabular}
	\label{tab:DMC_RL}
    \vspace{-10pt}
\end{table*}

\subsection{Continual DeepMind Control}
We run the continual learning procedure with $3$ seeds and report the mean rewards and standard deviations of $10$ episodes. 
As shown in Table~\ref{tab:DMC_RL}, our method achieves significant improvements compared with existing model-based and model-free RL approaches on all tasks. 
For example, our model performs nearly $2.5\times$ higher than DreamerV2 on the first learning task \textit{Walk}, and the performance on the task \textit{Downhill} is almost $7\times$ higher than CURL.
Additionally, our model generally outperforms the straightforward combinations of EWC and popular RL approaches. 
For each combination, we present the results using two different sets of hyper-parameters termed ``hyper-1'' to ``hyper-4''. 
For DV2+EWC, EWC coefficients are ($\omega_1=10^9$, $\omega_2=10^7$, $\omega_3=10^8$) for ``hyper-1'' and ($10^7$,$10^5$,$10^6$) for ``hyper-2'', where $\omega_1$, $\omega_2$, $\omega_3$ correspond to actor, value and world model losses respectively. For CURL+EWC, ``hyper-3'' uses ($10^6$,$10^3$,$10^7$) while ``hyper-4'' employs ($10^5$,$10^2$,$10^6$), with $\omega_3$ now regulates contrastive loss.
We can observe that the performance may change sharply when using different hyper-parameter sets. 
The reason may be that additional constraints on the feature extraction module from visual inputs make it difficult to preserve pre-learned knowledge in its parameters.
Furthermore, our approach outperforms the results of both multi-task joint training and single-task training on three of four tasks. Considering the average rewards on all tasks, our model outperforms the single training results by a large margin ($811.5$ vs. $741.3$). It shows that our approach not only mitigates catastrophic forgetting issues but also improves the pre-learned tasks.

\begin{figure*}[t]
\centering
\includegraphics[width=\linewidth]{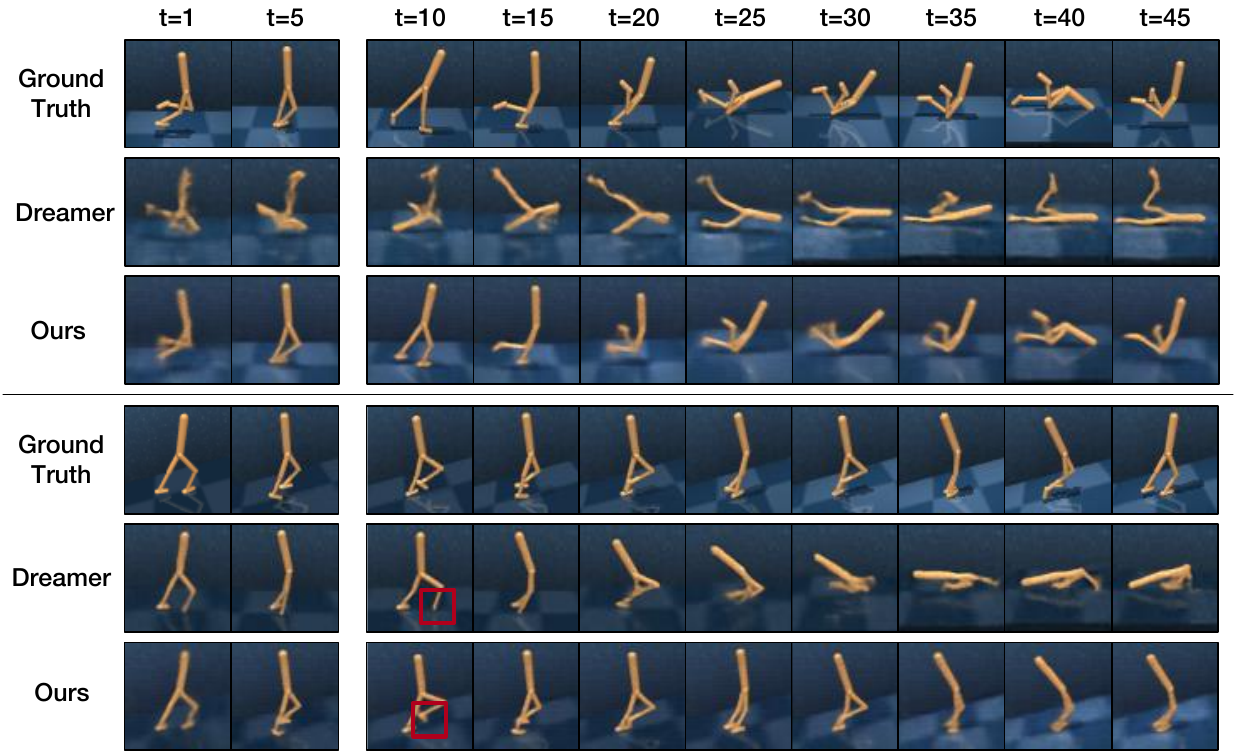}
\caption{Visual forecasting results on the task \textit{Walk} (\textbf{upper}) and \textit{Uphill} (\textbf{bottom}) after the model trained on the last task \textit{Nofoot}. For each sequence, we use the first $5$ images as context frames and predict the next $45$ frames given actions. The world model in our method effectively alleviates catastrophic forgetting, while DreamerV2 makes predictions similar to those in the last task.}
\label{fig:DMC_pred}
\end{figure*}

We also conduct visual forecasting experiments on the DMC benchmark to explore whether the learned world model can preserve the pre-learned visual dynamics during the continual learning procedure.
After training the models on the last task \textit{Nofoot}, we randomly collect sequences of frames and actions during the test phases of the first two tasks \textit{Walk} and \textit{Uphill}.
We input the first five frames to the learned world model and ask it to predict the next $45$ frames with actions input.
The prediction results are shown in Fig.~\ref{fig:DMC_pred}.
From these prediction results, we can observe that although these models have been continually trained on different tasks, the world model in our method still remembers the visual dynamics in previous tasks.
It successfully forecasts future frames given the corresponding action inputs without forgetting.
On the contrary, the results predicted by DreamerV2 suffer from severe blur effects and share a similar appearance with trajectories of the last task \textit{Nofoot}. 
These results show that the world models in MBRL methods can not handle catastrophic forgetting.

\begin{table*}[t]
\centering
    \caption{Results on Meta-World: The task order is \textit{Window-open} $\rightarrow$ \textit{Button-press} $\rightarrow$ \textit{Hammer} $\rightarrow$ \textit{Assembly}. We evaluate the models on all tasks after completing training on the last task (\textit{Assembly}).
	}
    \vspace{4pt}
\small
\setlength\tabcolsep{4pt}
    \begin{tabular}{lccccc}
    \toprule
   Method & Window-open & Button-press & Hammer & Assembly & Average  \\
    \midrule
    DV2 &220 $\pm$ 33  & 195 $\pm$ 232 & 331 $\pm $196   &1075 $\pm$ 37  & 451   \\
    CURL~\cite{DBLP:conf/icml/LaskinSA20}  & 152 $\pm$ 37 & 51 $\pm$ 9  & 491 $\pm$ 38    & 862 $\pm$ 151   & 389    \\
    DV2+EWC &549 $\pm$ 990 &267 $\pm$ 105 & 448 $\pm$ 55 &206 $\pm$ 10  & 367\\
    CURL+EWC   & 462 $\pm$ 450 & 265 $\pm$ 157 & 480 $\pm$ 41     & 202 $\pm$ 17  &352  \\
    Ours & \textbf{1004} $\pm$ 1251  & \textbf{681} $\pm$ 546 & \textbf{524} $\pm$ 333     & \textbf{2389} $\pm$ 676 & \textbf{1149} \\
    \midrule
    Multi-task  & 2119 $\pm$ 1528  & 2722 $\pm$ 516 &  2499 $\pm$ 895    & 2310 $\pm$ 418 & 2412 \\
    Single-task  & 3367 $\pm$ 954  & 2642 $\pm$ 840  & 950 $\pm$ 109    & 1283 $\pm$ 66 & 2060 \\
    \bottomrule
    \end{tabular}
    \vspace{-5pt}
	\label{tab:MW_RL}
\end{table*}

\subsection{Continual Meta-World} 
We can also find similar observations on the Meta-World benchmark in Table~\ref{tab:MW_RL}. Our approach consistently outperforms previous RL methods and their direct combinations with the continual learning EWC method on all tasks. 
For instance, our model achieves near $3\times$ average rewards compared with DreamerV2 and about $2.5\times$ average rewards compared with CURL. 
Notably, the result on the final task \textit{Assembly} shows that our method can effectively use the pre-learned knowledge to further improve the performance on the latter tasks. 
For the \textit{Assembly} task, our model outperforms the single training results by a large margin and even outperforms the joint training results. Combined with the results in the DMC benchmark, it seems that our framework has a huge potential in improving the forward transfer performance in continual visual control.

\subsection{Ablation Studies}
\label{sec:ablation}

\begin{table}[t]
\small
\centering
\caption{Ablation studies of each component in our model on the average rewards across all tasks in the DMC benchmark. 
``Replay'' denotes the use of the experience replay mechanism. 
``Action shuffling'' indicates using action shuffling in \newcontrib{}. ``Value regulation'' means that the agent takes the previous value model to constrain the value estimation.}
\vspace{4pt}
\setlength\tabcolsep{10pt}
\begin{tabular}{ccccc}
\toprule
Replay & $\epsilon$-Greedy & Action shuffling & Value regulation & Reward\\
\midrule
\no & \no & \no & \no  & 450 \\
\yes & \no & \no & \no  & 525  \\
\yes & \yes & \no & \no  & 616 \\
\yes & \yes & \yes & \no & 788 \\
\yes & \yes & \yes & \yes & 812 \\
\bottomrule
\end{tabular}
\label{tab:Ablations_RL}
\vspace{-5pt}
\end{table}

\begin{figure}[t]
\centering
\includegraphics[width=0.7\linewidth]{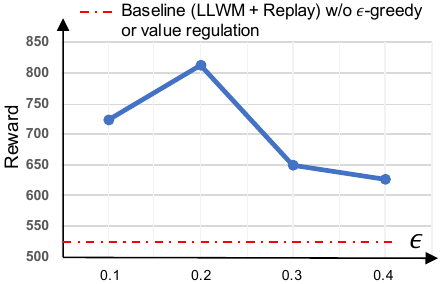}
\caption{Ablation study of different $\epsilon$ in the $\epsilon$-greedy exploration on DMC.}
\label{fig:ab_epsilon}
\end{figure}

We conduct ablation studies on the DMC benchmark to evaluate different components on the average rewards across all tasks. In Table~\ref{tab:Ablations_RL}, the first row provides the results of the baseline model, which only contains the life-long world model. 
In the second row, we train the baseline model with the experience replay mechanism, and we observe that the improvements are limited.
As suggested in \sect{sect:dul_reg}, the main reason lies in the fact that training on the replay data only will result in overfitting problems on both the reward prediction and the value estimation.
Then, as shown in the third row and the fourth row, we further improve the experience replay with the exploration-conservative behavior learning scheme.
We gradually introduce the $\epsilon$-greedy exploration strategy and the data augmentation on the frame-action pairs and use them to regularize the learning of the reward module on replay trajectories.
We can observe significant improvement compared with the pure replay scheme.
Finally, in the bottom row, we reuse the pre-learned value model to introduce the constraint on the value estimation during the behavior learning stage.
It also shows a large performance boost.
We also experiment with different $\epsilon$ values in the $\epsilon$-greedy exploration and show the average rewards in Fig.~\ref{fig:ab_epsilon}. On one hand, slightly increasing the probability to introduce random actions can significantly improve the performance. On the other hand, excessively raising the $\epsilon$ value may introduce too much noise and degenerate the performance.

Finally, we evaluate our continual MBRL approach over $3$ random task orders on the DMC benchmark to analyze whether our model can effectively alleviate catastrophic forgetting regardless of the task order. The mean reward with standard deviation is $801 \pm 35$, which shows that the proposed techniques, including the life-long world model and the experience replay mechanism, are still effective despite the change of training order. 

\section{Conclusion and Discussion}

In this paper, we studied the continual learning problem of sequential visual control tasks, which is challenging due to the dynamics shift.
%
The main contributions of our method can be viewed in two aspects.
First, it presents the life-long world model that captures task-specific visual dynamics in a Gaussian mixture latent space, incorporating an experience replay mechanism to overcome the forgetting issue in the world model.
Second, it further presents a pilot model-based RL approach for continual visual control by involving the \newcontrib{} scheme to overcome the overfitting problems within both the world model and the policy models. 
Our approach has shown competitive results on both DeepMind Control and Meta-World benchmarks, achieving remarkable improvements over the straightforward combinations of existing continual learning and reinforcement learning approaches.

In the future, extending the framework to handle more drastic domain shifts or out-of-distribution tasks encountered in lifelong learning scenarios would enhance its robustness and applicability. Investigating mechanisms that improve transfer learning and mitigate catastrophic forgetting across highly dissimilar tasks is a key direction.

\begin{credits}

\subsubsection{\discintname}

The authors have no competing interests to declare that are relevant to the content of this article.
\end{credits}

\section*{Acknowledgments}

This work was supported by the National Natural Science Foundation of China (62250062), the Smart Grid National Science and Technology Major Project (2024ZD0801200), the Shanghai Municipal Science and Technology Major Project (2021SHZDZX0102), and the Fundamental Research Funds for the Central Universities.


\begin{thebibliography}{10}
\providecommand{\url}[1]{\texttt{#1}}
\providecommand{\urlprefix}{URL }
\providecommand{\doi}[1]{https://doi.org/#1}

\bibitem{ammar2014online}
Ammar, H.B., Eaton, E., Ruvolo, P., Taylor, M.: Online multi-task learning for policy gradient methods. In: ICML. pp. 1206--1214 (2014)

\bibitem{DBLP:conf/iclr/BharadhwajBEL22}
Bharadhwaj, H., Babaeizadeh, M., Erhan, D., Levine, S.: Information prioritization through empowerment in visual model-based {RL}. In: ICLR (2022)

\bibitem{burchi2025learning}
Burchi, M., Timofte, R.: Learning transformer-based world models with contrastive predictive coding. In: ICLR (2025)

\bibitem{chen2022continual}
Chen, G., Zhang, W., Lu, H., Gao, S., Wang, Y., Long, M., Yang, X.: Continual predictive learning from videos. In: CVPR. pp. 10728--10737 (2022)

\bibitem{daniels2022model}
Daniels, Z.A., Raghavan, A., Hostetler, J., Rahman, A., Sur, I., Piacentino, M., Divakaran, A., Corizzo, R., Faber, K., Japkowicz, N., et~al.: Model-free generative replay for lifelong reinforcement learning: Application to starcraft-2. In: Conference on Lifelong Learning Agents. pp. 1120--1145 (2022)

\bibitem{dilokthanakul2016deep}
Dilokthanakul, N., Mediano, P.A., Garnelo, M., Lee, M.C., Salimbeni, H., Arulkumaran, K., Shanahan, M.: Deep unsupervised clustering with gaussian mixture variational autoencoders. arXiv preprint arXiv:1611.02648  (2016)

\bibitem{ha2018recurrent}
Ha, D., Schmidhuber, J.: Recurrent world models facilitate policy evolution. In: NeurIPS. vol.~31 (2018)

\bibitem{DBLP:conf/icml/HaarnojaZAL18}
Haarnoja, T., Zhou, A., Abbeel, P., Levine, S.: Soft actor-critic: Off-policy maximum entropy deep reinforcement learning with a stochastic actor. In: ICML. vol.~80, pp. 1856--1865 (2018)

\bibitem{hadsell2020embracing}
Hadsell, R., Rao, D., Rusu, A.A., Pascanu, R.: Embracing change: Continual learning in deep neural networks. Trends in cognitive sciences  \textbf{24}(12),  1028--1040 (2020)

\bibitem{hafner2020dream}
Hafner, D., Lillicrap, T., Ba, J., Norouzi, M.: Dream to control: Learning behaviors by latent imagination. In: ICLR (2020)

\bibitem{hafner2019learning}
Hafner, D., Lillicrap, T., Fischer, I., Villegas, R., Ha, D., Lee, H., Davidson, J.: Learning latent dynamics for planning from pixels. In: ICML. pp. 2555--2565 (2019)

\bibitem{hafnermastering}
Hafner, D., Lillicrap, T.P., Norouzi, M., Ba, J.: Mastering atari with discrete world models. In: ICLR (2021)

\bibitem{hafner2023mastering}
Hafner, D., Pasukonis, J., Ba, J., Lillicrap, T.: Mastering diverse domains through world models. arXiv preprint arXiv:2301.04104  (2023)

\bibitem{hansen2022modem}
Hansen, N., Lin, Y., Su, H., Wang, X., Kumar, V., Rajeswaran, A.: Modem: Accelerating visual model-based reinforcement learning with demonstrations. arXiv preprint arXiv:2212.05698  (2022)

\bibitem{hinton2015distilling}
Hinton, G., Vinyals, O., Dean, J.: Distilling the knowledge in a neural network. arXiv preprint arXiv:1503.02531  (2015)

\bibitem{igl2021transient}
Igl, M., Farquhar, G., Luketina, J., B{\"o}hmer, J., Whiteson, S.: Transient non-stationarity and generalisation in deep reinforcement learning. In: ICLR (2021)

\bibitem{jiang2017variational}
Jiang, Z., Zheng, Y., Tan, H., Tang, B., Zhou, H.: Variational deep embedding: An unsupervised and generative approach to clustering (2017)

\bibitem{khetarpal2022towards}
Khetarpal, K., Riemer, M., Rish, I., Precup, D.: Towards continual reinforcement learning: A review and perspectives. Journal of Artificial Intelligence Research  \textbf{75},  1401--1476 (2022)

\bibitem{kirkpatrick2017overcoming}
Kirkpatrick, J., Pascanu, R., Rabinowitz, N., Veness, J., Desjardins, G., Rusu, A.A., Milan, K., Quan, J., Ramalho, T., Grabska-Barwinska, A., et~al.: Overcoming catastrophic forgetting in neural networks. Proceedings of the national academy of sciences  \textbf{114}(13),  3521--3526 (2017)

\bibitem{kostrikov2020image}
Kostrikov, I., Yarats, D., Fergus, R.: Image augmentation is all you need: Regularizing deep reinforcement learning from pixels. arXiv preprint arXiv:2004.13649  (2020)

\bibitem{DBLP:conf/nips/KumarZTL20}
Kumar, A., Zhou, A., Tucker, G., Levine, S.: Conservative q-learning for offline reinforcement learning. In: NeurIPS (2020)

\bibitem{lan2022memory}
Lan, Q., Pan, Y., Luo, J., Mahmood, A.R.: Memory-efficient reinforcement learning with knowledge consolidation. arXiv preprint arXiv:2205.10868  (2022)

\bibitem{DBLP:conf/icml/LaskinSA20}
Laskin, M., Srinivas, A., Abbeel, P.: {CURL:} contrastive unsupervised representations for reinforcement learning. In: ICML. pp. 5639--5650 (2020)

\bibitem{li2025open}
Li, J., Wang, Q., Wang, Y., Jin, X., Li, Y., Zeng, W., Yang, X.: Open-world reinforcement learning over long short-term imagination. In: ICLR (2025)

\bibitem{liotet2022lifelong}
Liotet, P., Vidaich, F., Metelli, A.M., Restelli, M.: Lifelong hyper-policy optimization with multiple importance sampling regularization. In: AAAI. pp. 7525--7533. No.~7 (2022)

\bibitem{liu2021conflict}
Liu, B., Liu, X., Jin, X., Stone, P., Liu, Q.: Conflict-averse gradient descent for multi-task learning. In: NeurIPS. pp. 18878--18890 (2021)

\bibitem{mendez2022reuse}
Mendez, J.A., Eaton, E.: How to reuse and compose knowledge for a lifetime of tasks: A survey on continual learning and functional composition. arXiv preprint arXiv:2207.07730  (2022)

\bibitem{mendez2022modular}
Mendez, J.A., van Seijen, H., Eaton, E.: Modular lifelong reinforcement learning via neural composition. arXiv preprint arXiv:2207.00429  (2022)

\bibitem{paniso}
Pan, M., Zhu, X., Wang, Y., Yang, X.: Iso-dream: Isolating and leveraging noncontrollable visual dynamics in world models. In: NeurIPS (2022)

\bibitem{powers2022cora}
Powers, S., Xing, E., Kolve, E., Mottaghi, R., Gupta, A.: Cora: Benchmarks, baselines, and metrics as a platform for continual reinforcement learning agents. In: Conference on Lifelong Learning Agents. pp. 705--743 (2022)

\bibitem{rao2019continual}
Rao, D., Visin, F., Rusu, A.A., Teh, Y.W., Pascanu, R., Hadsell, R.: Continual unsupervised representation learning. In: NeurIPS (2019)

\bibitem{riemer2019scalable}
Riemer, M., Klinger, T., Bouneffouf, D., Franceschini, M.: Scalable recollections for continual lifelong learning. In: AAAI. vol.~33, pp. 1352--1359 (2019)

\bibitem{sekar2020planning}
Sekar, R., Rybkin, O., Daniilidis, K., Abbeel, P., Hafner, D., Pathak, D.: Planning to explore via self-supervised world models. In: ICML. pp. 8583--8592 (2020)

\bibitem{shi2021meta}
Shi, G., Azizzadenesheli, K., O'Connell, M., Chung, S.J., Yue, Y.: Meta-adaptive nonlinear control: Theory and algorithms. In: NeurIPS. pp. 10013--10025 (2021)

\bibitem{sun2024learning}
Sun, R., Zang, H., Li, X., Islam, R.: Learning latent dynamic robust representations for world models. In: ICML (2024)

\bibitem{tassa2018deepmind}
Tassa, Y., Doron, Y., Muldal, A., Erez, T., Li, Y., Casas, D.d.L., Budden, D., Abdolmaleki, A., Merel, J., Lefrancq, A., et~al.: Deepmind control suite. arXiv preprint arXiv:1801.00690  (2018)

\bibitem{DBLP:conf/nips/WolczykZPKM21}
Wolczyk, M., Zajac, M., Pascanu, R., Kucinski, L., Milos, P.: Continual world: {A} robotic benchmark for continual reinforcement learning. In: NeurIPS. pp. 28496--28510 (2021)

\bibitem{xie2022lifelong}
Xie, A., Finn, C.: Lifelong robotic reinforcement learning by retaining experiences. In: Conference on Lifelong Learning Agents. pp. 838--855 (2022)

\bibitem{yarats2021improving}
Yarats, D., Zhang, A., Kostrikov, I., Amos, B., Pineau, J., Fergus, R.: Improving sample efficiency in model-free reinforcement learning from images. In: AAAI. vol.~35, pp. 10674--10681 (2021)

\bibitem{yu2020gradient}
Yu, T., Kumar, S., Gupta, A., Levine, S., Hausman, K., Finn, C.: Gradient surgery for multi-task learning. In: NeurIPS. pp. 5824--5836 (2020)

\bibitem{DBLP:conf/corl/YuQHJHFL19}
Yu, T., Quillen, D., He, Z., Julian, R., Hausman, K., Finn, C., Levine, S.: Meta-world: {A} benchmark and evaluation for multi-task and meta reinforcement learning. In: CoRL. vol.~100, pp. 1094--1100 (2019)

\bibitem{zhang2022catastrophic}
Zhang, T., Wang, X., Liang, B., Yuan, B.: Catastrophic interference in reinforcement learning: A solution based on context division and knowledge distillation. IEEE Transactions on Neural Networks and Learning Systems  (2022)

\bibitem{zhou2022forgetting}
Zhou, W., Bohez, S., Humplik, J., Heess, N., Abdolmaleki, A., Rao, D., Wulfmeier, M., Haarnoja, T.: Forgetting and imbalance in robot lifelong learning with off-policy data. In: Conference on Lifelong Learning Agents. pp. 294--309 (2022)

\end{thebibliography}
\end{document}